# OBESEYE: Interpretable Diet Recommender for Obesity Management using Machine Learning and Explainable AI

Mrinmoy Roy[1*], Srabonti Das[2], Anica Tasnim Protity[3]

[1]Data Analyst, Department of Clinical Informatics, Ascension, Chicago, United States of America
[2]Nutritionist, Al Helal Specialized Hospital Ltd., Dhaka, Bangladesh
[3]Student, Department of Biological Sciences, Northern Illinois University, Dekalb, United States of America

*Abstract*: Obesity, the leading cause of many non-communicable diseases, occurs mainly for eating more than our body requirements and lack of proper activity. So, being healthy requires heathy diet plans, especially for patients with comorbidities. But it is difficult to figure out the exact quantity of each nutrient because nutrients requirement varies based on physical and disease conditions. In our study we proposed a novel machine learning based system to predict the amount of nutrients one individual requires for being healthy. We applied different machine learning algorithms: linear regression, support vector machine (SVM), decision tree, random forest, XGBoost, LightGBM on fluid and 3 other major micronutrients: carbohydrate, protein, fat consumption prediction. We achieved high accuracy with low root mean square error (RMSE) by using linear regression in fluid prediction, random forest in carbohydrate prediction and LightGBM in protein and fat prediction. We believe our diet recommender system, OBESEYE, is the only of its kind which recommends diet with the consideration of comorbidities and physical conditions and promote encouragement to get rid of obesity.

## 1. Introduction

Diet planning involves making conscious decisions based on the type and amount of food and drink. Food monitoring helps patients understand food intake for energy production, tissue repair, active immune system, well-organized cognitive function, and healthy bowel movement. A healthy diet can significantly support optimal health and minimize the risk of chronic illnesses like heart disease, diabetes, obesity, chronic kidney disease, and hypertension. Appropriate monitoring of food intake is vital for patients as this enables medical practitioners to evaluate nutritional intake, maximize medication effectiveness, detect eating triggers, and ensure dietary recommendations. To better understand the relationship between their eating habits and disease progression, the patient can record their daily food consumption, which trained dietitians can further evaluate and suggest. For example, avoiding calorie and fat-dense food, limiting salt, sugar, and fluid intake, and ensuring proper nutrition by a prescribed nutritionist are essential for patients with non-communicable diseases (NCDs), especially diabetes, heart disease, chronic kidney disease, and hypertension. Although NCDs are not fully recovered, these diseases can be controlled. Maintaining a healthy lifestyle by changing food habits can pertain to longevity and quality of life. Controlled food uptake helps to manage the patient's body weight and other parameters, such as sugar level, cholesterol, and serum creatinine, which further aids in managing the overall NCD-related health risks. This food monitoring approach helps patients and primary physicians alleviate disease symptoms, improve health conditions, and support the treatment process.

An adequate nutritional balance can be achieved by tracking the consumption of liquids, proteins, carbohydrates, and fats. Food intake based on these four different nutritional categories can be adjusted with the extent of single or combined diseases. For instance, people with protein deficiency might need more protein than usual, whereas patients with chronic kidney diseases might require less protein. Similarly, people with cardiovascular diseases may benefit from reducing their intake of saturated and trans fats.

Nowadays, AI-based software and cellular applications are becoming more critical in healthcare services for the early detection of cancers and other skin-related diseases [1], prediction of genetic disorders, analysis of medical reports, and the covid-related awareness. Machine learning is becoming increasingly popular in the field of patient food recommendations. Because of its remarkable ability to analyze large amounts of data, machine learning can identify patterns and correlations that can be used to make personalized food charts. Using medical records, dietary habits, and food preferences, machine learning algorithms can accurately generate food recommendations. Some food-log websites and mobile apps that use machine learning to recommend a healthy meal plan by analyzing the user's medical records, dietary needs, and food preferences are: Eat this much, Nutritionix, and Diabeedoc.

In developed countries, registered dietitians are available for patients in almost every medical center. Besides, AI-based food prediction applications are used by cellphone users to fulfill their diet goals. This mainly involves determining the regular calorie intake, tracking exercise, and suggesting which food should be eaten or prohibited. However, access to a prescribed nutritionist is usually limited or unavailable in developing countries. Moreover, the food logs used for most cellphone-based applications use Western foods, which only partially reflect the actual calorie count in Oriental-style recipes. Besides, recommendations from AI-based software might not work for all patients because the same food can exhibit different physiological effects on other individuals.

---
*Corresponding author: mrinmoy.cs10@gmail.com

In this study, we used a machine learning based approach to predict the appropriate amount of nutrients based on the individual's physical and disease related information. We developed a estimation system that can predict the food intake for individuals with NCDs. We have different ML models with the food log from 146 NCD individuals. After training, analyzing, and test-training the models, we developed a system that can successfully predict the recommended food chart for fluid and three other major micronutrients: carbohydrate, protein, fat using the processed data. This machine learning prediction approach could help healthcare professionals and patients with NCDs manage their food intake, therefore maintaining a healthy lifestyle.

## 2. Related Works

The impact of food intake on the body and mind is significant for both healthy and unhealthy individuals. In particular, hospitalized patients can greatly benefit from a robust diet recommended by a dietician. Such a diet can increase longevity, protect against disease, improve quality of life, and reduce detrimental outcomes like readmission, extended length of stay in hospitals, mortality, and hospital costs [2], [3]. Iwendi el al. proposed deep learning solutions based on disease and physical features and achieved 97.74% accuracy using LSTM model. The study only used data from 30 patients but didn't discuss overfitting problems. Again, medical personnel are yet to understand the function of the recommender system and creates the need of explanation of the model using explainable AI which is missing in this study. Nowadays, evaluating food intake has become a crucial part of scientific research and clinical trials because the same diet can have different effects on individuals or population groups due to variations in demographics and health conditions [4]. Capturing the complex relationship between nutrition and disease is challenging due to numerous influencing factors [5].

However, with the availability of extensive food data and machine learning algorithms, data-driven models can now provide a comprehensive view considering all variables. Diet planning application using teacher-forced REINFORCE algorithm generates controllable sequence from composition patterns of diet data and construct database of menus and diets [11]. Currently, screening tools in hospitals use naive machine learning algorithms like K-means to classify inpatients based on their nutrition status [5]. This decision system also assists in healthcare staff management and automates approaches. Machine learning algorithms correlate patient socioeconomic, demographic, and health characteristics, as well as institutional factors, with their food intake [2]. Various ML algorithms, such as logistic regression, decision trees, random forests, recurrent neural networks, gated recurrent units, and long short-term memory, are being employed in diet recommendation systems. Deep learning classifiers enhance precision and accuracy by incorporating k-clique [6].

Precision nutrition, a new trend in nutrition research, focuses on understanding individual DNA, microbiome, and metabolic responses to specific dietary plans. This approach is becoming popular for selecting effective eating plans to prevent or treat diseases [7]. Precision nutrition using ML provides patient-level monitoring and identifies metabolomic signature differences to establish the causal relationship between food intake and individuals' metabolic responses. Additionally, ML is applied to assess true dietary intake using image, video, and sensor data from devices like wristwatches or mobile phones, rather than relying on estimated intake [5]. Patient self-management applications deploying complex ML algorithms benefit both patients and healthcare professionals. In diabetic management, continuous remote monitoring of symptoms and biomarkers results in improved glycemic control, including reductions in fasting and postprandial glucose levels, glucose excursions, and glycosylated hemoglobin [8].

Patients often need to follow specific diet plans before surgery or therapy to avoid complications and enhance recovery after surgery (ERAS). Implementing ERAS protocols significantly reduces the length of hospital stays (LHOS) and readmissions in colorectal surgeries, gastrointestinal surgery, and other fields [9]. The conventional strategy of prolonged fasting can lead to excessive catabolism, resulting in weight loss and muscle mass loss. ML algorithms can be employed in ERAS to generate a personalized diet plan for the weeks following surgery, such as radical cystectomy with or without neobladder, based on patient responses [10]. Additionally, current developments in the field of explainable AI (EXAI) are incorporating SHAP, LIME, and different ML algorithms to provide patient and healthcare practitioners satisfaction by explaining the reasons for selecting specific foods.

## 3. Data Collection

Based on the study conducted over three months from October 2022 to December 2022, we investigated the prevalence of non-communicable diseases (NCDs) in 146 patients of both genders, aged between 18 and 95 years, admitted to a hospital in the Northern part of Dhaka, Bangladesh. Collective data on the patient's demographic information, such as age, sex, height, weight, waist and hip measurements, and vital parameters, including blood sugar, serum creatinine, blood pressure, the presence or absence of NCDs, and the amount of food uptake suggested by the nutritionist were taken accordingly. Among the NCDs observed in the study population were diabetes mellitus, chronic kidney disease, cardiovascular disease, chronic respiratory illness, irritable bowel syndrome, and thyroid dysfunction. Based on the patient's medical records and body statistics, a nutritionist then prescribed the amount of carbohydrates, protein, fat, and water content. The water quantity was measured in units of liters (L). Although cups are preferred for measuring food intake, this was done to provide more accurate and consistent results. A cup of cooked white rice might weigh 150 or 200 grams for measuring food intake, depending on the cup size and the rice variety, which cannot accurately track food intake. Therefore, our study measured the quantity in grams (g) for the solid food categories.

In addition to the reasons mentioned above, nutritionists may also use grams of rice to measure food intake because it is a more precise unit of measurement. Cups can vary in size, so measuring the amount of rice in a cup can be difficult. On the other hand, grams are a more precise unit of measurement, and they can be easily measured using a kitchen scale, making it easier for nutritionists to track the exact amount of rice a person consumes.

## 4. Data Description

The variables included in this study are patients' demographic and physical information, along with their underlying diseases. We gathered data on each patient's gender, age, height, weight, and waist-hip ratio. Height was measured in centimeters, weight in kilograms, and waist-hip ratio in inches. For the DM variable, fasting blood glucose test values below 5.7 mmol/L and above 3.9 mmol/L were considered normal and coded as "No." Values above 5.7 mmol/L were classified as hyperglycemia and coded as "High," while values below 3.9 mmol/L were considered hypoglycemia and coded as "Low." To detect CKD, we used the serum creatinine value. A creatinine level greater than 1.2 mg/dL for women and greater than 1.4 mg/dL for adult men was considered indicative of CKD [12]. We reviewed the patients' medical history to identify any heart-related diseases, marking "Yes" for heart problems and "No" otherwise.

Patient blood pressure was measured using a sphygmomanometer due to its higher accuracy compared to digital devices [13]. Blood pressure was measured in millimeters of mercury (mmHg). Diastolic blood pressure below 60 mmHg and systolic blood pressure below 90 mmHg [36] were categorized as hypotension and marked as "Low." Diastolic pressure between 70-80 mmHg and systolic pressure between 90-120 mmHg were classified as "Normal," while values higher than these ranges were considered hypertension, indicated by "High." IBS, respiratory illness, and thyroid issues were recorded based on patients' experiences, previous medical history, and continuous monitoring. If any of these diseases were present, "Yes" was recorded under the corresponding disease, and "No" otherwise. Among the dependent variables fluid is numeric numbers with two decimal places and measured in liter. Other dependent variables carbohydrate, protein and fat are numeric type and measured in grams.

## 5. System Workflow

The overall system workflow is shown in fig. 1 where at the beginning of the execution, the patient's physical characteristics data and NCDs related data are combined to get the raw training data. After getting the training data, preprocessing step starts and converts the data into required formats. The preprocessed data is then fed to machine learning models for prediction tasks. The linear regression machine learning algorithm is used for fluid requirements prediction in liter, random forest algorithm is used in carbohydrate prediction in grams and LightGBM is used to predict protein and fat requirements in grams.

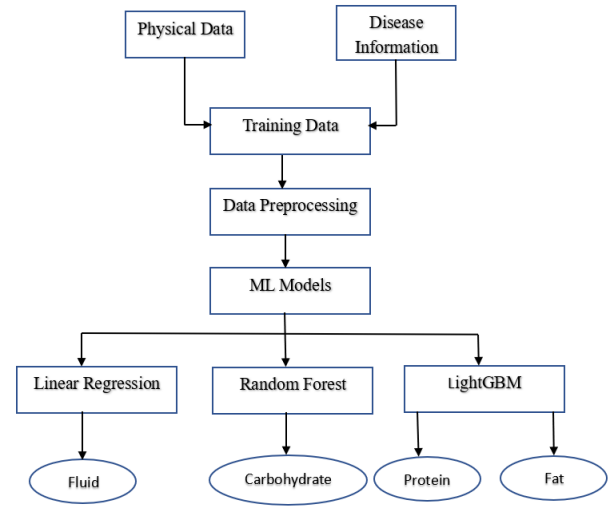
Fig. 1. Diet recommender workflows

## 6. Proposed Model

### A. Data Preprocessing

We have collected 146 patients physical and disease related information with nutritionist provided dieting chart for the three main macronutrients: carbohydrates, protein, fat, and daily required fluid consumption. Patients' physical information height is converted to cm from inch-feet measurement for feeding the data into ML models. We categorized diabetes values into two groups: being diabetic as "Yes" and not being diabetic as "No". The same categorization process goes for the creatinine values. Again, patients based on their systolic and diastolic pressure divided into three categories: low pressure, normal pressure, and high pressure. Diastolic pressure less than 60 or systolic pressure less than 90 is considered as low pressure, diastolic pressure in between 60 to 80 and systolic pressure in between 90 to 120 is considered as normal pressure whereas more than these range for both diastolic and systolic is considered as high blood pressure.

Moreover, patients with no fluid consumption limitation are given specific values for machine learning models prediction. According to [14], men with no fluid consumption limitation are encoded as 2.6 liter whereas women are encoded as 2 liters. Patients are marked -1 as sugar level value or creatinine values in case of having no diabetes or CKD. All other categorical features having binary values are encoded using binary encoding technique: 0 for "no" and 1 for "yes". Gender is encoded as 0 for "female" and 1 for "male". Blood pressure also encoded using nominal encoding technique: 0 for "low", 1 for "normal" and 2 for "high". All the variables are typecast from object type based on ML algorithms input and output criteria. We also divided the age variable into 3 corresponding groups: "<=35" for age equal to or less than 35 years, "35<>65" for age in between 35 and 65 years, and ">=65" for more than 65 years old age.

### B. Machine Learning Model

Our aim is to predict 3 main micronutrients: carbohydrates, protein, fat in grams and daily fluid requirement in liter. So, the prediction task is a regression problem and we applied 6 different machine learning regression algorithms for the appropriate number prediction. First, we divided the dataset into 80-20 train-test-split. Then we trained the training dataset using different ML algorithms and tested the performance using test dataset. We leveraged linear regression, support vector machine (SVM), decision tree, random forest, XGBoost, lightGBM algorithms for each target variable prediction and compared their performance using R square, root mean square error (RMSE) and overall accuracy. For SVM, we used linear kernel type with regularization parameter C as 2.0. For random forest regression analysis, we took 100 number of trees in the forest.

Additionally, in the case of XGBoost, we limited the number of nodes in the tree to 3, number of boosting stages to 500, learning rate as 0.1 and gbtree as the booster. For lightGBM, we used default gradient boosting decision tree, maximum number of leaves in each tree 10 and learning rate 0.05. We also leveraged the random forest feature importance to find the most important features in case of each target variable. We tested these 6 types of algorithms because of their easy implementation, faster training speed, better accuracy, and lower memory usage. We decided on the final model after comparing the model's performance based on RMSE and then applied grid search cross-validation to calculate the score in different combination of parameters for each final model. This grid search technique improves the final model performance of carbohydrate and protein prediction substantially.

### 7. Results

We compared the ML models based on RMSE because it measures the average difference between the predicted value and actual value and provides the estimation about the model performance in target value prediction [15]. Another advantage is RMSE has the same unit as the target variable which makes it easy to interpret. We calculated R square value as it indicates the variance percentage of the target variable in the collective response of independent variables and how well the model can fit the underlying data [16]. We also estimated the model overall test accuracy using the mean of percentage difference of each predicted value from the test value which presents the model performance in terms of unknown data. The model performances for fluid prediction in terms of RMSE, R square and accuracy are shown in table 1.

We selected the linear regression model for fluid prediction as it predicts the fluid value with a small average difference of 0.39 liter and has high accuracy. The tree-based algorithm decision tree shows overfitting behavior and has an R square value of 1.00. Linear regression fluid prediction test values with corresponding predicted value are shown in fig. 2.

Table 1
Fluid prediction metrics

| Algorithm | RMSE | $R^2$ Value | Accuracy (%) |
|---|---|---|---|
| Linear | 0.39 | 0.31 | 78.75 |
| SVM | 0.41 | 0.25 | 79.58 |
| DS | 0.53 | 1.00 | 73.89 |
| RF | 0.41 | 0.85 | 75.45 |
| XGBoost | 0.48 | 0.99 | 72.22 |
| LightGBM | 0.39 | 0.27 | 75.83 |

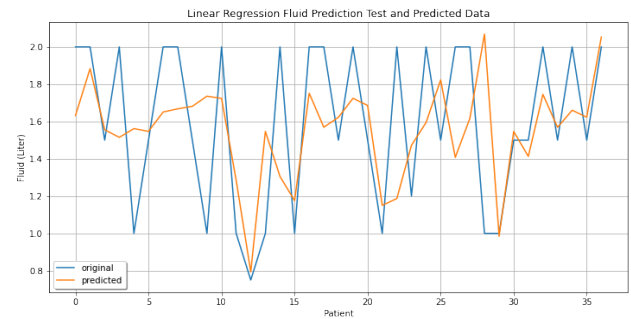

Fig. 2. Linear regression fluid prediction value vs. actual value

Random forest regression algorithm shows the best performance in carbohydrate prediction. All the metrics for carbohydrate prediction are shown in table 2.

Table 2
Carbohydrate prediction metrics

| Algorithm | RMSE | $R^2$ Value | Accuracy (%) |
|---|---|---|---|
| Linear | 35.61 | 0.52 | 83.61 |
| SVM | 31.79 | 0.42 | 85.45 |
| DS | 38.49 | 1.00 | 83.34 |
| RF | 29.08 | 0.89 | 86.99 |
| XGBoost | 36.08 | 0.99 | 82.86 |
| LightGBM | 31.70 | 0.43 | 86.02 |

Among the models, random forest has the lowest RMSE 29.08 with 82.86% accuracy. This RMSE score is comparatively small in comparison to the 102 grams to 260 grams range and tolerable according to the nutritionist for specific disease conditions. Random forest carbohydrate prediction with actual test data is shown in fig. 3.

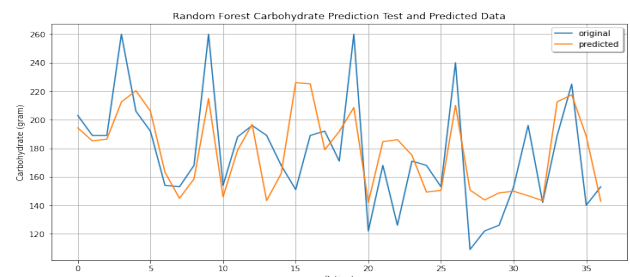

Fig. 3. Random forest carbohydrate prediction value vs actual value

Machine learning models' metrics for protein prediction are shown in table 3. where LightGBM shows the least RMSE 15.95 with accuracy 79.27%.

Table 3
Protein prediction metrics

| Algorithm | RMSE | R² Value | Accuracy (%) |
|---|---|---|---|
| Linear | 17.05 | 0.4257 | 78.33 |
| SVM | 18.78 | 0.33 | 75.96 |
| DS | 20.28 | 1.00 | 77.69 |
| RF | 16.05 | 0.90 | 79.49 |
| XGBoost | 17.79 | 0.99 | 77.50 |
| LightGBM | 15.95 | 0.55 | 79.27 |

LightGBM model's test data prediction with actual data is shown in fig. 4.

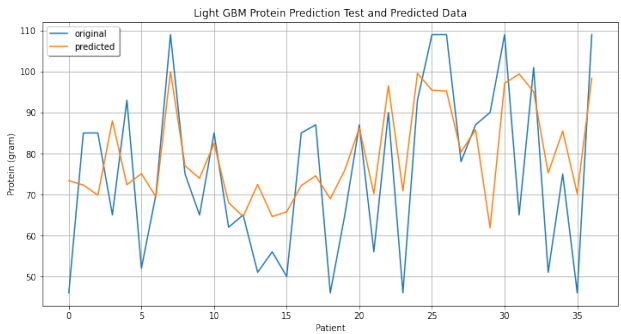

Fig. 4. LightGBM protein prediction value vs. actual value

Again, lightGBM model gives the most optimal performance in case of fat prediction which is shown in table 4. Though RMSE 15.09 is not very good in comparison to the patients 25 grams to 88 grams range, according to nutritionist it is tolerable with the diet calculation's margin of error consideration. LightGBM fat prediction values with actual test data is presented in fig. 5.

Table 4
Fat prediction metrics

| Algorithm | RMSE | R² Value | Accuracy (%) |
|---|---|---|---|
| Linear | 16.15 | 0.46 | 65.72 |
| SVM | 14.31 | 0.38 | 70.82 |
| DS | 16.25 | 1.00 | 79.66 |
| RF | 15.37 | 0.89 | 68.36 |
| XGBoost | 18.35 | 0.99 | 63.89 |
| LightGBM | 15.09 | 0.43 | 64.96 |

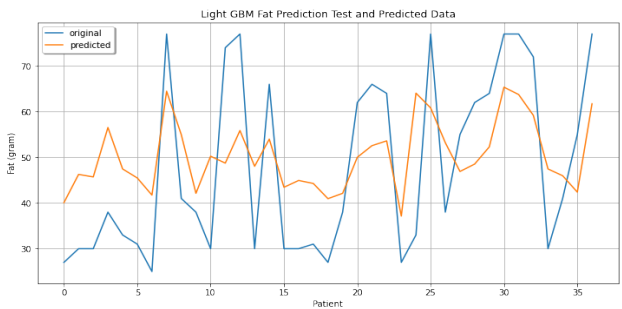

Fig. 5. LightGBM fat prediction value vs. actual value

## 8. Model Interpretability

Today's machine learning models are taking major decisions on behalf of human. So, it is very crucial to trust the black box artificial intelligence models completely and the trust only begins if we can understand the logic behind the model's decision. Therefore, nowadays every machine learning model should be easily interpretable by their user. As our proposed diet recommender model's decisions need to be accepted by a wider range of patients, we applied explainable AI techniques to properly explain our model's decisions. First, we described our model's individual decision using a local interpretable model LIME (Local Interpretable Model-Agnostic Explanations) which is shown in fig. 6.

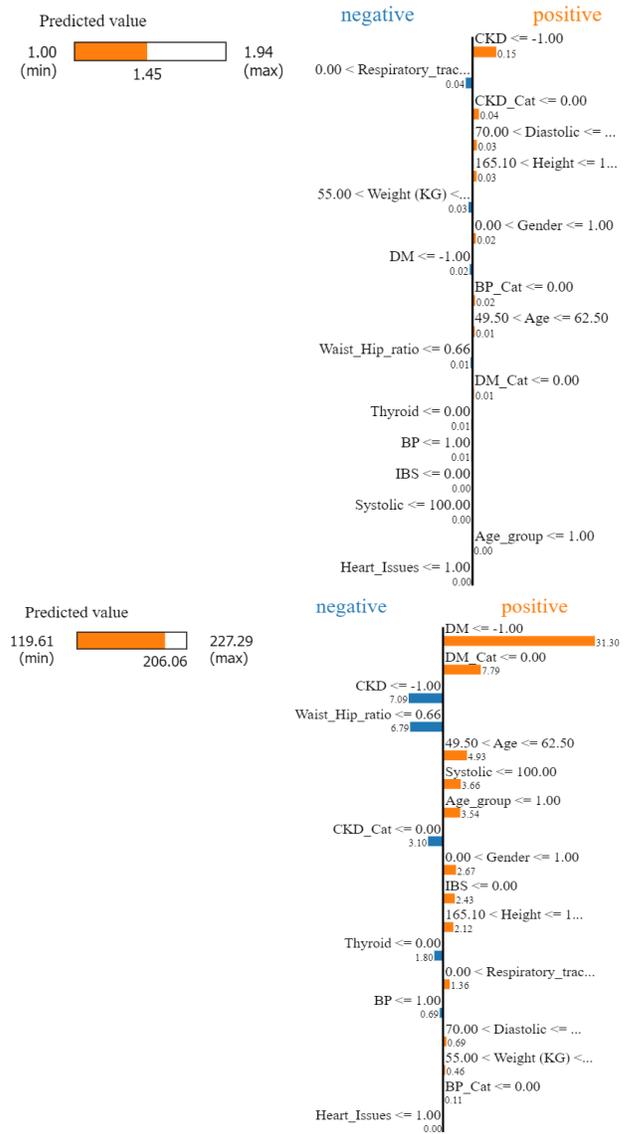

Fig. 6. Fluid (top) and Carbohydrate (bottom) prediction local explanation by LIME

The first part of fluid prediction shows the predicted value of 1.45 liter for one sample and second part describes that CKD, height, gender, BP categorization (BP_Cat), age groups have positive influence and RTI, weight, waist hip ratio have negative influence on predicted value. Same goes for the carbohydrate prediction where predicted value is 206.06 grams and DM, DM categorization (DM_Cat), systolic blood pressure, age group, gender have positive influence whereas CKD, CKD categorization, thyroid and waist hip ratio have negative influence. LIME explanation for protein and fat diet predictions are shown in fig. 7 with the predicted values and independent features positive and negative influence on them.

We also applied SHAP (Shapley Additive Explanations) for global interpretation of our model. SHAP summary plot which is shown in fig. 8 combines feature importance with feature effects. In the case of overall fluid prediction, CKD is the most important feature whereas in fig. 9 DM is for carbohydrate prediction. Moreover, the X-axis presents the distribution of the shapely values for each feature from low to high.

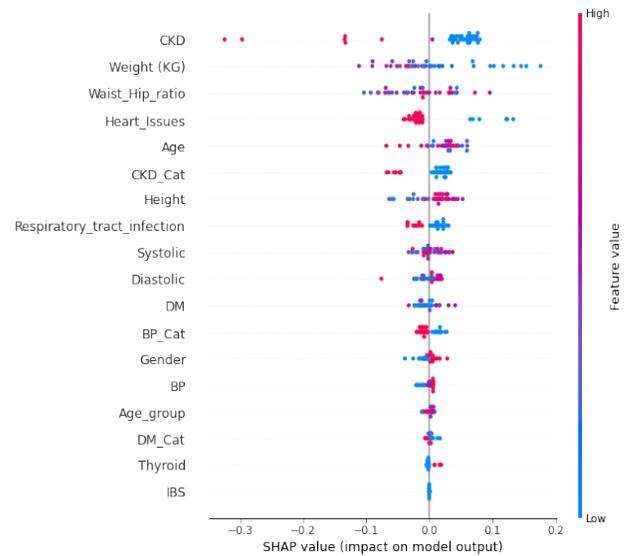

Fig. 8. Fluid prediction global explanation by SHAP

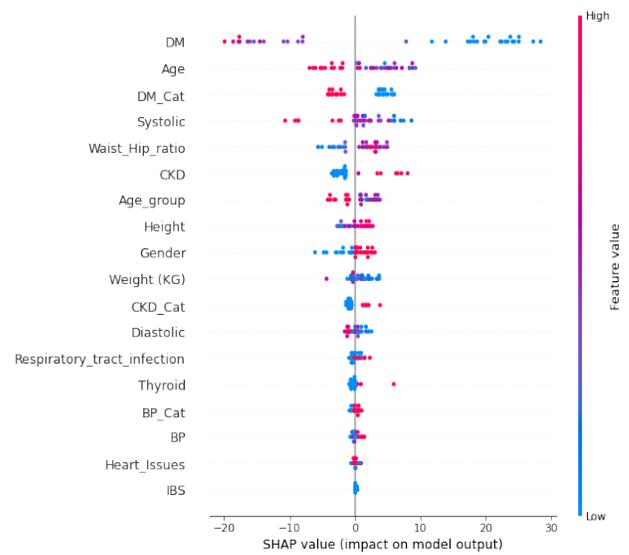

Fig. 9. Carbohydrate prediction global explanation by SHAP

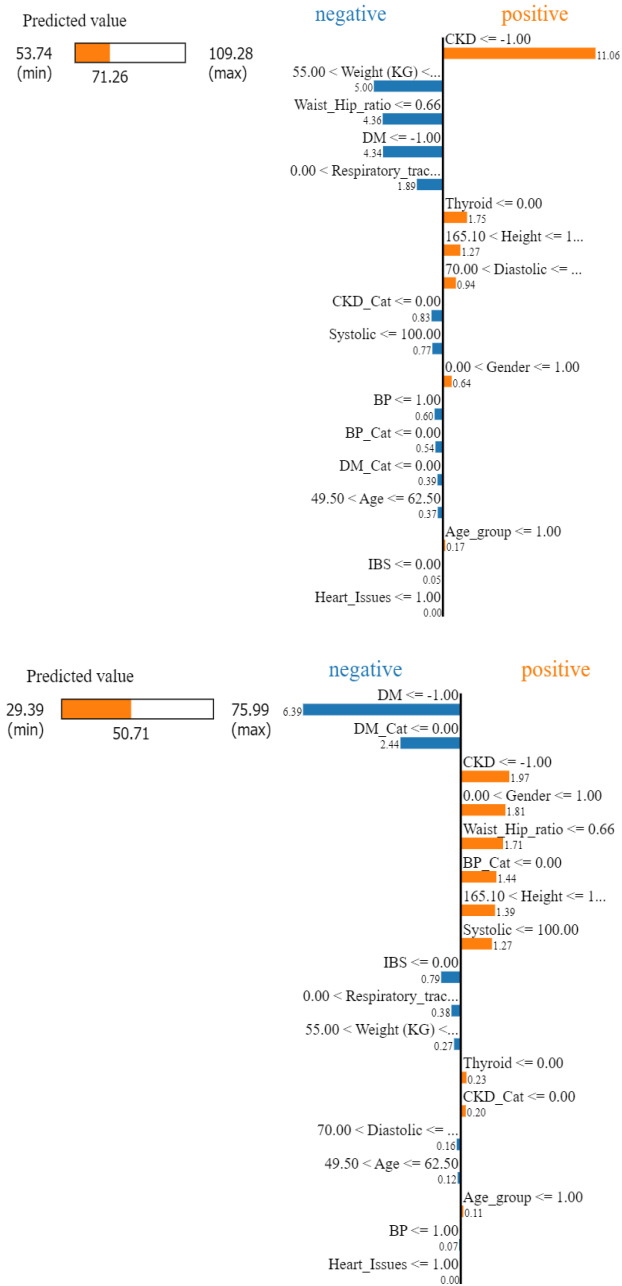

Fig. 7. Protein (top) and Fat (bottom) prediction local explanation by LIME

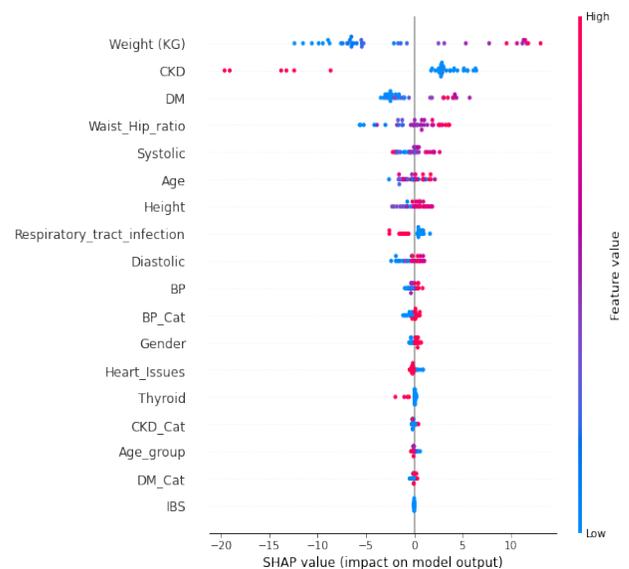

Fig. 10. Protein prediction global explanation by SHAP

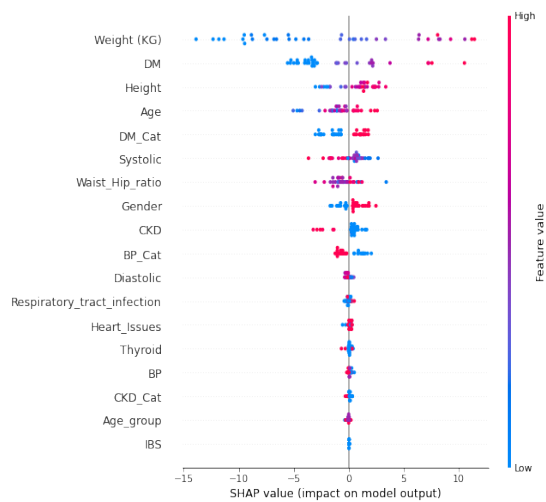

Fig. 11. Fat prediction global explanation by SHAP

## 9. Limitations and Future Works

Machine learning models can predict the correct values based on training stage. This stage needs lots of data to analyze the different combinations of features. Though we collected patients' information from different age groups, height, weight and gender, the final model's prediction can be improved if we incorporate more patient information in our training. Moreover, we recommended a diet for only 3 major micronutrients and fluid consumption. In future, we want to improve our model performance by collecting more data from patients to reflect the overall scenario. We also want to add other demographic information: economic, transportation, housing, other vulnerabilities for accurate diet prediction. Based on disease conditions and side effects, our future research will recommend specific food items to patients and extend our diet recommendation to other important food nutrients like vitamins, minerals etc.

## 10. Conclusion

Food consumption than required generates more calories which increases the risk of getting obese. At the same time, it makes people vulnerable to many non-communicable diseases. In the US 41.9% of adults and 19.7% children are affected by obesity [17] which in turn increases the risk of high blood pressure, heart diseases, kidney diseases, diabetics [18]. This is even more difficult for patients with comorbidities, even can kill a person slowly. To avoid the scenario, the importance of a diet recommender is irrefutable. Our proposed novel machine learning based diet recommender system used linear regression for fluid, random forest for carbohydrate and LightGBM for protein and fat requirements prediction because of their low RMSE values. By leveraging our diet recommender system, people can restrain themselves from overeating and maintain a healthy lifestyle.

## References

[1] M. Roy and A. T. Protity, "Hair and Scalp Disease Detection using Machine Learning and Image Processing", in *COMPUTE*, vol. 3, no. 1, pp. 7–13, Jan. 2023.

[2] L. J. Curtis, R. Valaitis, C. Laur, T. McNicholl, R. Nasser, and H. Keller, "Low food intake in hospital: patient, institutional, and clinical factors," in *Applied Physiology, Nutrition, and Metabolism*, vol. 43, no. 12, pp. 1239–1246, Dec. 2018.

[3] C. Iwendi, S. Khan, J. H. Anajemba, A. K. Bashir, and F. Noor, "Realizing an Efficient IoMT-Assisted Patient Diet Recommendation System Through Machine Learning Model," in *IEEE Access*, vol. 8, pp. 28462–28474, 2020.

[4] L. Oliveira Chaves, A. L. Gomes Domingos, D. Louzada Fernandes, F. Ribeiro Cerqueira, R. Siqueira-Batista, and J. Bressan, "Applicability of machine learning techniques in food intake assessment: A systematic review," in *Critical Reviews in Food Science and Nutrition*, pp. 1–18, Jul. 2021.

[5] D. Kirk, E. Kok, M. Tufano, B. Tekinerdogan, E. J. M. Feskens, and G. Camps, "Machine Learning in Nutrition Research," in *Advances in Nutrition*, vol. 13, no. 6, pp. 2573–2589, Nov. 2022.

[6] S. Manoharan and Satish, "Patient diet recommendation system using K clique and deep learning classifiers," in *Journal of Artificial Intelligence and Capsule Networks*, vol. 2, no. 2, pp. 121-130,

[7] 677 H. Avenue, Boston, and Ma 02115, "Precision Nutrition," The Nutrition Source, Sep. 23, 2020. https://www.hsph.harvard.edu/nutritionsource/precision-nutrition/#:~:text=Precision%20nutrition%20evaluates%20one (accessed Jun. 01, 2023).

[8] S. Ellahham, "Artificial Intelligence: The Future for Diabetes Care," in *The American Journal of Medicine*, vol. 133, no. 8, pp. 895-900, Apr. 2020.

[9] H. C. Yi et al., "Impact of Enhanced Recovery after Surgery with Preoperative Whey Protein-Infused Carbohydrate Loading and Postoperative Early Oral Feeding among Surgical Gynecologic Cancer Patients: An Open-Labelled Randomized Controlled Trial," in *Nutrients*, vol. 12, no. 1, pp. 264, Jan. 2020.

[10] M. Melnyk, R. G. Casey, P. Black, and A. J. Koupparis, "Enhanced recovery after surgery (ERAS) protocols: Time to change practice?," in *Canadian Urological Association Journal*, vol. 5, no. 5, pp. 342–348, Oct. 2011.

[11] C. Lee, S. Kim, C. Lim, J. Kim, Y. Kim, and M. Jung, "Diet Planning with Machine Learning: Teacher-forced REINFORCE for Composition Compliance with Nutrition Enhancement," in *KDD '21: Proceedings of the 27th ACM SIGKDD Conference on Knowledge Discovery & Data Mining*, pp. 3150-3160, 2021.

[12] National Kidney Foundation. "Tests to Measure Kidney Function, Damage and Detect Abnormalities," in *National Kidney Foundation*, 6 Nov. 2017, www.kidney.org/atoz/content/kidneytests

[13] B. Shahbabu et al., "Which is More Accurate in Measuring the Blood Pressure? A Digital or an Aneroid Sphygmomanometer," in *Journal of clinical and diagnostic research*, vol. 10, no. 3, pp. LC11-LC14, March 2016.

[14] Better Health Channel, "Water - A vital nutrient," Vic.gov.au, Sep. 11, 2021. https://www.betterhealth.vic.gov.au/health/healthyliving/water-a-vital-nutrient

[15] "SAP Help Portal," help.sap.com. https://help.sap.com/docs/SAP_PREDICTIVE_ANALYTICS/41d1a6d4e7574e32b815f1cc87c00fd42/5e5198fd4afe4ae5b48fefe0d3161810.html

[16] J. Frost, "How to Interpret R-squared in Regression Analysis," Statistics by Jim, 2018. https://statisticsbyjim.com/regression/interpret-r-squared-regression/

[17] CDC, "Notice of Funding Opportunity: HOP," Centers for Disease Control and Prevention, Feb. 01, 2023. https://www.cdc.gov/nccdphp/dnpao/state-local-programs/fundingopp/2023/hop.html#:~:text=Obesity%20in%20the%20United%20States

[18] Centers for Disease Control and Prevention, "The Health Effects of Overweight and Obesity," Centers for Disease Control and Prevention, Sep. 24, 2022. https://www.cdc.gov/healthyweight/effects/index.html